# A Semantic Analyzer for the Comprehension of the Spontaneous Arabic Speech

Mourad Mars, Mounir Zrigui, Mohamed Belgacem, Anis Zouaghi

{Mourad.mars, Mohamed.belgacem}@e.u-grenoble3.fr
Mounir.zrigui@fsm.rnu.tn
Anis.zouaghi@riadi.rnu.tn

**Abstract.** This work is part of a large research project entitled "Oréodule" aimed at developing tools for automatic speech recognition, translation, and synthesis for Arabic language. Our attention has mainly been focused on an attempt to improve the probabilistic model on which our semantic decoder is based. To achieve this goal, we have decided to test the influence of the pertinent context use, and of the contextual data integration of different types, on the effectiveness of the semantic decoder. The findings are quite satisfactory.

## 1. Introduction

Our work fits within the framework of the automatic comprehension of Arabic speech. The use of statistical models for the speech recognition and comprehension [1] [2], have the merit to strongly reduce the resort to human expertise. They can also be applied to other fields such as multilingual applications [3].

The automatic association with each word of the recognized utterance of the proper group of FSe (Semantic Feature) [4] based on such models generally requires the analysis of the context. The contextual information plays a major role in the selection of the adequate FSe. These pieces of information spare the trouble of interpretation ambiguities and improve the performance of the comprehension system of [6] [10].

In the standard approach, the decoding of the word meaning is generated by analysing the context which precedes it or/and which follows it immediately. However, in the case of the comprehension of spontaneous Arabic, this is not always optimum. Indeed, we have recorded rather high error rates sometimes equal to 57% and 48,6%. The first rate results of analysing the meaning of the word preceding the target word, and the second rate by using the meaning of the two preceding words (see figure 3, paragraph 5). In order to sort out this problem, we have decided not to take into account the meaning of the pertinent words to select the target word sense. We have also taken into account the type of illocutionary act achieved by the utterance to which belongs the word to be interpreted (refusal, request, etc.) and of its nature (for example request for reservation, timing, etc.) for the prediction of FSe to be used.

## 2. The Difficulties of the Semantic Decoding of the Spontaneous Arabic Speech

### 2.1. Varieties of the Spoken Arabic Language

Arabic is the sixth most widely spoken language in the world with approximately 250 million speakers. For historical and ideological reasons, this language presents a hierarchy of varieties:

- The modern standard Arabic: It is the written language of literature and press, witch is usually spoken on the radio, in conferences and official speeches all over the Arab countries. Standard Arabic is learned at school.

- The intermediate Arabic: it is a simplified alternative of standard Arabic and the well shaped of the dialectical Arabic. It borrows its lexicon from the dialect as well as standard Arabic. Currently this alternative is increasing. It is more and more frequently used in the studies and the medias. It allows approaching the illiterate and the native language of the people.

- The dialectical Arabic: it is another alternative of the classical Arabic. It is mainly oral; it is the language of daily conversation. Each Arab country has its own dialect. Although there are several dialects, mutual comprehension is possible between the different countries.

These various registers make the automatic processing of the Arabic language impossible with its several varieties. That's why the developed systems are conceived to process only one of these alternatives. The semantic decoding model proposed in this article is dedicated to the modern standard Arabic language, since it is the official language of all the Arabic countries.

### 2.2. Particularities of Arabic Language

The automatic comprehension of the natural language is a very hard task. The major difficulties related to the automatic processing of the spontaneous Arabic speech are:

- The non-vowelization of the majority of the Arabic texts in books and newspapers makes the training task when using a probabilistic model more complicated. At the semantic level, the automatic detection of the sense of a unvowelized word is very ambiguous. For example, the word مدرسة can have three possible interpretations according to its vowelization. It may mean a school or a teacher (female) or taught (past participle of teach). This problem is similar to the ambiguity resulting from homonyms in other languages.

- An Arabic word may express a whole French or English expression [8]. For example the word أرأيت expresses in English ″Have you seen″. Thus the automatic interpretation of such words requires their preliminary segmentation, which is not an easy task.

- The connection without space of the coordinating conjunction و "and" to the words. It is rather hard to distinguish between the و, as a letter of a word (for example

ورق "sheets") and the و having the role of a coordinating conjunction (utterance E). However this type of conjunction plays a significant role in the recognized utterance interpretation, by identifying its proposals.

(E) أريد معرفة توقيت القطار الذاهب إلى تونس وحجز مكان.

(I would like to know the departure time of the train going to Tunis and booking a place.)

- The order followed to arrange the words in an utterance, is variable. This complicates the construction task of an adapted language model, from which the recognized utterance will be interpreted.

- The possibility of existence of several graphemic realisations for same phoneme, or several phonetic realisations for same grapheme. Even some graphemes can't be considered during the pronunciation [9]. This aspect makes the recognition harder.

- Some letters of Arabic language for instance: ظ - ذ - ض – خ – ح – ف are pronounced by using strong expiration, so the quality of the microphone can affect the recognition of the speech results.

- The essence of pronunciation of some letters as for example: غ – ر.

## 3. The Proposed Approach for the Semantic Decoding of Arabic Speech

### 3.1. The Conventional Methods Used

In literature several methods have been proposed for semantic analysing of the spontaneous speech. Some use the HMM (such as [3], [8], [6]), others the neuronal networks ([9], [10]), the n-grams language models ([11], [12]), the λ-calculus ([13]), or also logics ([14]). The table 1 below shows the main formalisms used for the language comprehension, their advantages and disadvantages (The list is not exhaustive owing to the lack of space).

Unlike Latin language, the processing of spontaneous Arabic speech remains, without sufficient consideration by scientific research. During the last two decades the efforts were rather concentrated on the realization of the morphological and syntactic analyzers for Arabic ([15], [16], [20]etc). In spite of the importance of the representation and of the semantic analysis for the realization of any comprehension system, there are only some works in this field which are interested in the processing of Arabic language (such as [17], [18], [19], [21]). Al Biruni system [21] for example, is based on a combination of the Fillmore case grammar formalism and of the Mel'cuk sense-text theory, for the semantic analysis, the representation of the Arab text and the handling of its representation. As for [17], he uses the unification grammar HDPSG of [22] which allows the integration of syntactic and semantic knowledge in the same grammar, in order to lead to a deep analysis. All these works are interested in the processing of the written Arabic rather than the spoken one. The method that we are proposing on this paper is inspired of the grammar case. It allows decoding the words meaning of the recognized utterance while being based on the relevant contextual data, i.e. by considering only the context which has a semantic influence on the word to be interpreted. The advantage of our method is that the context is automatically determined and needs no human expert intervention. Moreover, the contextual data which contribute to a word interpretation are of several types: semantic, illocutionary, and linguistic. The consideration of different types of contextual data has enabled us to improve the performance of our decoder. Moreover, our model is adapted to the processing of spoken Arabic, since it is based only on the analysis of the elements carrying meanings in the speaker's request. The insignificant or redundant elements are ignored.

**Table 1.** Examples of formalisms used for language comprehension.

| Used formalisms | Examples | Efficient for processing | Advantages | Disadvantages |
|---|---|---|---|---|
| HMM | [3], [6] , [8] | speech | Existence of powerful algorithms (such as Viterbi and A*) allowing to determine the optimal solution. | Requires corpus of bulky sizes. |
| Neuronal Networks | [9], [10] [18] | Writing and speech | Capacity of generalization and flexibility. | Coasty time development and very complex generated structures. |
| HDPSG | [17] | Writing | Allows an explicit integration in only one structure, the different linguistic analysis levels: phonetic, syntactic, and semantic. | Not adapted to be used in an interactive vocal system, (it allows analysing a sentence in term of syntactic component) |
| Case frames | [3] [21] | speech | Authorizes the treatment of the sentences without respecting grammatical rules and requires less expertise in linguistics. | Reduced the role of syntax. |

Following are the proposed approach for the semantic analysis of the spontaneous Arabic speech, and the way of the automatic extraction of the relevant context.

### 3.2. The Characteristics of our Approach

To device our semantic decoder, we have opted for the following choices:

- A componential representation of the meaning words: each significant word for the application is represented by a group of FSe = {field, semantic class, micro semantic dash} and a group of syntactic dashes Fsy = {gender, number, nature}. The dashes of the FSe group indicate respectively: the application field, the semantic class to which the word to be interpreted belongs, and the last dash is a micro semantic dash which helps to distinguish the meaning of the words belonging to the same semantic class. We note that the synonymic words or those having the same semantic role have the same group of FSe. By applying this representation, the meaning of the word "الذاهب" going for example is described as follows: الذاهب → FSe = {"transport"نقل, "movement" حركة, "destination"وجهة} + Fsy = {"masculine" مذكر, "singular" مفرد, "name" اسم}.

- A selective analysis: for the semantic decoding of the recognized utterance, we have relied rather on a semantic analysis and considered only the significant elements for the application. The blank words are eliminated during the pre-processing of the utterance phase, by using a lexical filter. This analysis is more tolerant as for the grammatical errors which characterize the spontaneous speech. Moreover, it doesn't need a high standard language of knowledge.

- A man/machine co-operation method based on corpus analysis (see figure 1): for building up our structure of sense representation SRS such as it is defined in [4], we developed a method based on a corpus analysis to extract significant words, reference words and semantic classes of the application, and on man-machine co-operation for word's interpretation. According to this method the user role is to indicate and to attribute the group of FSe and Fsy to words. And the machine role is to satisfy the constraint's integrities in order to lead to an unambiguous SRS. Our system is based on about ten constraints. An example of constraint to check is that: two different words can't be described by the same group of FSe only if they are considered as synonymic or having a same semantic role.

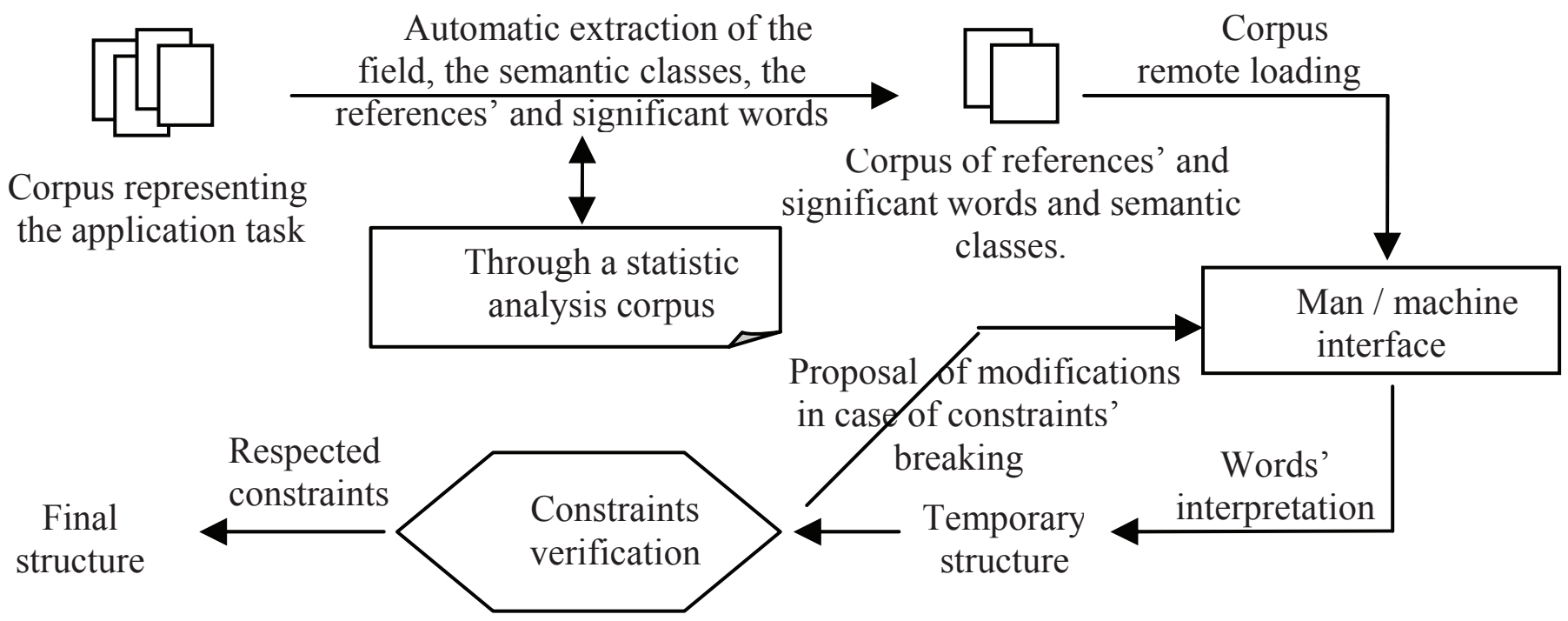

**Fig. 1.** The building steps of an application's SRS.

Notice that to extract the referring words (words indicating the type and the illocutionary essence of an utterance) we have adapted the TfxIdf method (Term frequency × Inverse document frequency), here after the formula:

$$p_{ij}=[tf\,(m_i,\,D_j).\log\,(n/df\,(m_i))]/[tf\,(m_i,\,D_j)+0{,}5+(1.5.n.l(D_j)/\sum_{Dk} l(Dk)).\log(n+1)]$$

Where, n and $l(D_k)$ stand respectively for, the number of considered requests' types and the length of all the requests relating to the corpus representing the field $D_k$ of the considered finalised application. The term tf $(m_i, D_j)$ indicates the number of occurrences of $m_i$ in $D_j$. df $(m_i)$ corresponds to the number of requests' types including $m_i$.

So the referring words associated to the requests of the type $D_j = \{mi \,/\, p_{ij} > \text{given threshold}\}$, where $p_{ij}$ indicates the weight of the word mi in the requests of $D_j$ type.

As for the extraction of semantic classes or the application concepts, we have used the k-means algorithm.

This defined method makes the task of words' interpretation much easier the same goes for the task of building up the SRS and keeping its coherence. The following figure 2 shows the man/machine interface through which each significant word is interpreted via the groups FSe and Fsy.

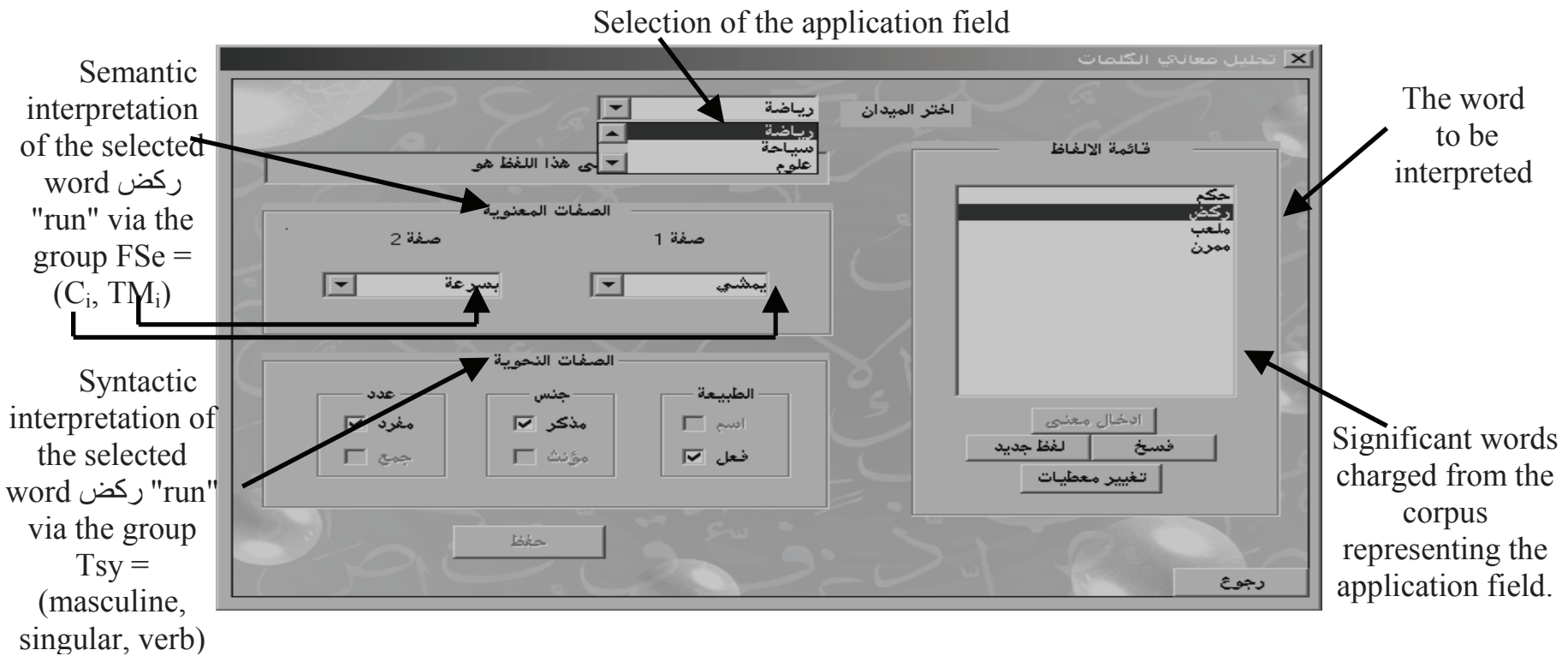

**Fig. 2.** The man/machine interface allowing the interpretation of significant words.

- A probabilistic grammar: this grammar contributes to choose the adequate FSe for the utterance significance description. It enables us to include several contextual data at the same time. Furthermore, it considers only the pertinent informations for the prediction of the word sense. The following equation stands for the interpretation probability of a word $M_i$ by the couple (Ci, TMi) taking into account the utterance type. We notice that in this formula we haven't considered the field application since it has been defined in advance. In our instance, it is about train information field.

$$P((C_i, TM_i) / M_i, NT_j) = P(NT_j / Mr_k) \times P(C_i / NT_j, M_{i-1}, CP_{i-1}, CP_{i-2}) \times P(TM_i / Ci, FSeP_{i-1}) \quad (1)$$

It is clear that this probability is expressed by terms of three conditional probabilities product. The first probability P ($NT_j$ / $Mr_k$) allows the identification of utterance type, if it is about a booking request, or cancelling a ticket, etc. Thus, taking into account words of references $Mr_k$ presents in the speaker's utterance. The words of references are unigram, or bi-grams, or trigrams (which can be separate) whose occurrence probabilities are equal to one. For example the bi-grams $_2$حجز $_1$أريد corresponding to 4-grams "I want to book" in english is a word of reference indicating that it is about booking request. This first probability is calculated as follows:

$$P(NTj / Mrk) = N(NTj(E), Mrk) / N(Mrk)$$

Where, N (NTj(E), Mrk) indicates the number of Mrk words' occurrence in the utterances of $NT_j$ type. And N(Mrk) is the total number of Mrk occurrences in the same utterance.

The second probability P ($C_i$ / $NT_j$, $M_{i-1}$, $CP_{i-1}$, $CP_{i-2}$) enables us to determine the semantic class $C_i$ to which belongs the word Mi, taking in account the utterance type and the two preceding pertinent semantic classes. This probability is calculated as follows:

$$P(C_i / NT_j, M_{i-1}, CPi_{-1}, CP_{i-2}) = N(NTj(E), C_i, CP_{i-1}, CP_{i-2}) / N(NTj(E), CP_{i-1}, CP_{i-2})$$

And the third probability P ($TM_i$ / $C_i$, $FSeP_{i-1}$) enables us to determine the micro semantic dash $TM_i$ to be attributed to Mi, while taking into account the class which has been attributed to this word and of preceding pertinent FSe. This last probability is calculated as follows:

$$P(TM_i / C_i, FSeP_{i-1}) = N(FSei, FSeP_{i-1}) / N(C_i, FSeP_{i-1})$$

In paragraph 4 is stated the method that we defined for the extraction of pertinent FSe.

### 3.3. The Semantic Analyzing Principle

We understand by the semantic analyzing of an utterance, the labelling of each one of its significant words through a group of FSe.

As shown in figure 3, the semantic decoding of the pre-processing utterance is based on the probabilistic language model of [23] and on semantic lexicon. The probabilistic model contributes to the selection of FSe to be affected to words of the utterance in order to be interpreted, and the semantic lexicon describes the meaning of each word through a group of FSe and a group of Fsy. It is from the decoded utterance that its meaning is deducted. This is by filling the attributes of the identified diagram with the corresponding values.

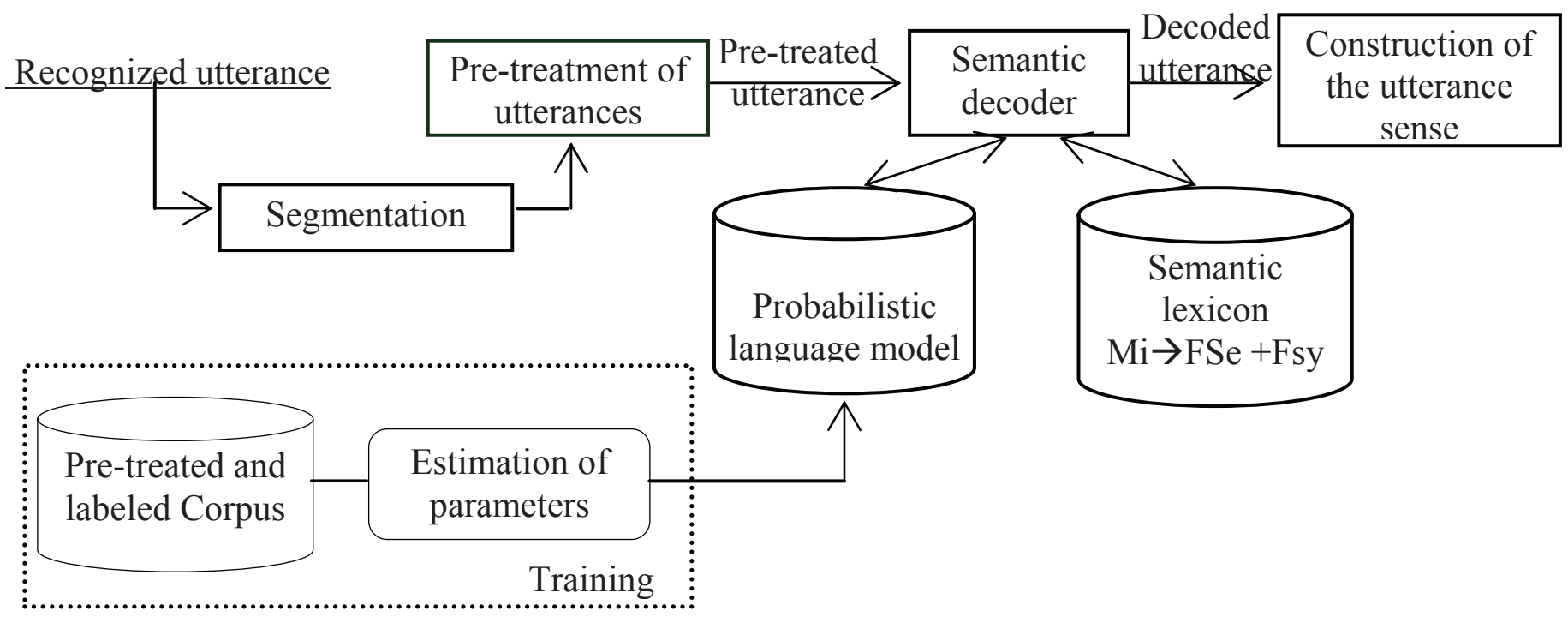

**Fig. 3.** The proposed semantic analyzer architecture.

During the training stage we considered a labelled and pre-processed corpus for the estimation of the probabilistic model parameters. The pre-processing of the representing corpus application enabled us to simplify the complexity and to reduce the size of the probabilistic model. This pre-processing like the pre-processing of the utterance consists in eliminating for example the blank words, by gathering certain words in only one entry, etc.

# 4. Extraction of the Pertinent Context for Semantic Decoding

## 4.1. The Extraction Principle

To determine the group of FSe to be assigned to the utterance words, we use in our probabilistic model the group of pertinent FSe only. We understand by pertinent FSe, the FSe used to describe words meaning which have a strong semantic affinity with the word Mi. Thus to identify the semantic class Ci to which the word Mi belongs, we consider in the equation E1 the two semantic classes $CP_{i-1}$ and $CP_{i-2}$ of the two FSe assigned to two words having the strongest semantic affinities with Mi. Similarly to determine the micro semantic dash TMi allowing the differentiation in the meaning of Mi of the other words of the same class Ci, we consider in E1 only the group FSePi = {CPi, TMPi } which was assigned to the word having the strongest semantic affinity with Mi. To achieve this goal, we have relied on the concept of average mutual information [24] which helps to calculate the correlation degree of two given words.

## 4.2. Calculation of the Semantic Affinity

Let's consider a recognized utterance E to be interpreted: $E = M_1 M_2 ... M_n$. Let $ME = \{ME_{-K} ..., ME_{-1}\ ME_1 \ldots, ME_K\}$ the group of words surrounding the word $M_i$ to be interpreted by considering a window of K size. As our model considers only the right context (see remark 1) of $M_i$ for the choice of FSe to be assigned to this word, the group ME is so reduced to $MEd = \{M_1, M_2 \ldots M_{i-1}\}$ (K is variable), which is the group of words preceding Mi. Now, to find the strongest semantic affinity between Mi and its context, we start by calculating the average mutual information between Mi and each of the words belonging to MEd.

Remark 1. *The Arabic language is written from right to left.*

Here below the formula of average mutual information IMm [23]:

$$IMm(Mi, MEdj) = P(Mi, MEdj) \times Log\, [P(Mi / MEdj) / P(Mi) \times P(MEdj)] + P(\overline{Mi}, \overline{MEdj}) \times Log\, [P(\overline{Mi} / \overline{MEdj}) / P(\overline{Mi}) \times P(\overline{MEdj})] + P(Mi, \overline{MEdj}) \times Log\, [P(Mi / \overline{MEdj}) / P(Mi) \times P(\overline{MEdj})] + P(\overline{Mi}, MEdj) \times Log\, [P(\overline{Mi} / MEdj) / P(\overline{Mi}) \times P(MEdj)]; \quad \text{with } 1 \leq j \leq i-1 \qquad \textbf{(2)}$$

We preferred to use IMm (equation 2) rather than traditional mutual information IM (equation 3), because the first measurement is more effective. Indeed IMm allows the calculation of the impact of the word absence on the appearance of the other.

$$IM(Mi, MEDj) = Log\, [P(Mi, MEdj) / P(Mi) \times P(MEDj)]; \quad \text{avec } 1 \leq j \leq i-1 \qquad \textbf{(3)}$$

The maximum semantic affinity *AffM* which the word Mi has with its right context is obtained by the following formula:

$$AffM(Mi, MEd) = argmax_j\, I(Mi, MEDj) = argmax_j\, Log[P(Mi, MEDj) / P(Mi) › P(MEDj)] \qquad \textbf{(4)}$$

We notice that we consider the FSe of the nearest word Mi, in the case of obtaining two equal semantic affinities.
Here is an illustrating example of words extraction having the biggest semantic affinity with the word to be interpreted تونس (Tunis: the capital of Tunisia) in the following request R:

ما هو سعر الذهاب إلى تونس. (word by word translation: What is the price going to Tunis)

Following a comparison of semantic affinities that the word Tunis has with each word of its right context (see figure 4), we notice that the words الذهاب (going) and إلى (to) have the biggest semantic affinity with تونس (Tunis). So, will be used the classes $CP_{i-1}$= مؤشر (mark) and $CP_{i-2}$= حركة (movement), which are attributed to the words الذهاب (going) and إلى (to) for the determination of the semantic class to which belongs the word Tunis, and the group $FSeP_{i-1}$={$CP_{i-1}$= مؤشر (mark), $TMP_{i-1}$ = وجهة (destination)} for the prediction of the micro semantic dash of the word Tunis.

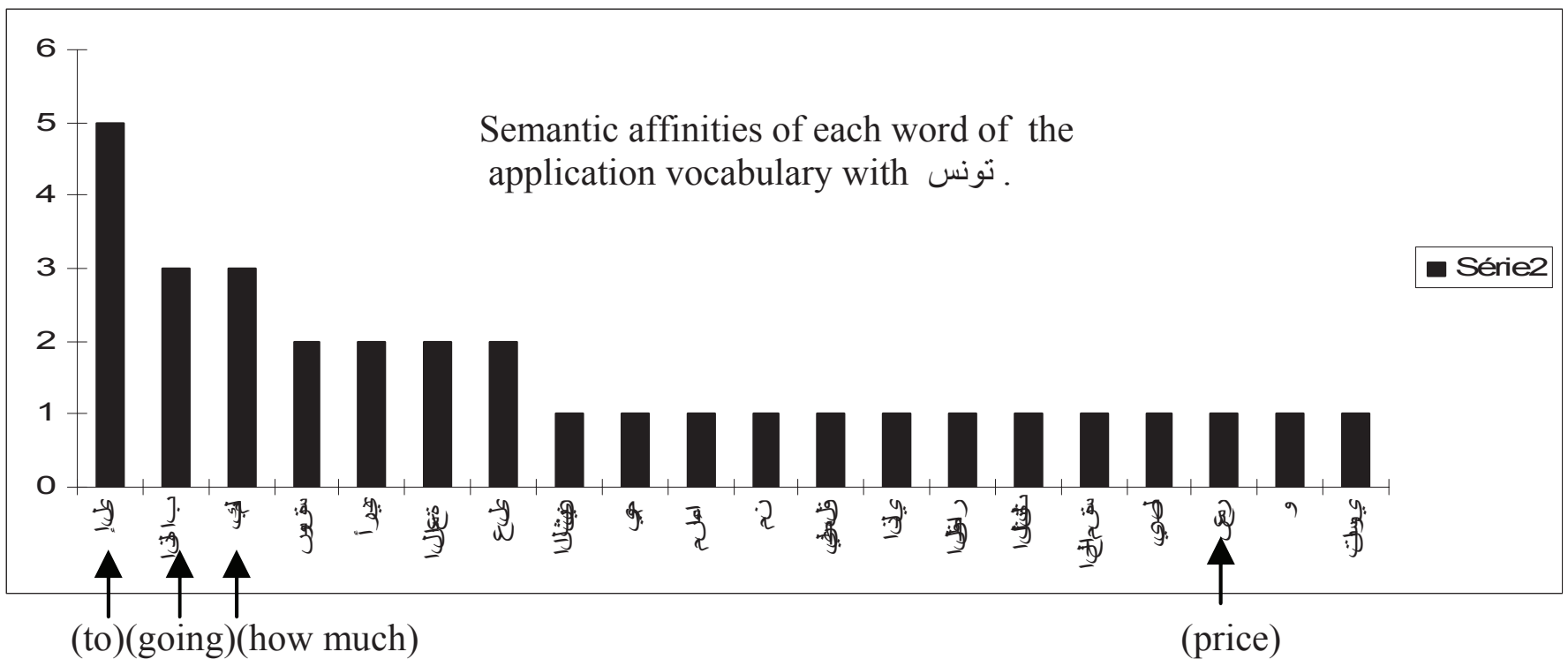

**Fig. 4.** Semantic affinities sample that the word تونس (Tunis) has with the other words of the application vocabulary.

In figure 4, the words indicated by an arrow are in fact those which have appeared in the right context of the word Tunis in the utterance (R) stated above.

## 5. Application of the Model and Results

We have used a hundred utterances different from those of the training corpus, carrying all on timing requests for the test. The training corpus (constituted of 10000 utterances representing the railway information field) was labelled with 37 FSe. To judge the quality of our decoder, we have calculated the percentage of FSe which are incorrectly assigned, using the following formula: $R_{error} = N_{inc} / N \times 100$.

Where, $N_{inc}$ is the number of FSe incorrectly assigned, and N is the total number of FSe assigned by an expert to the test corpus. N is equal to 500 in this test. The following figures 5 and 6 show respectively the consideration influence of:

- The pertinent context on the interpretation result with regard to models taking in account and in advance a fixed and determined historic.

- And the several types of contextual information and of the context length on the interpretation results. We notice that the use of semantic and illocutionary knowledge for parsing some foreign languages (such as English and French) has already been investigated, but not yet for arabic language.

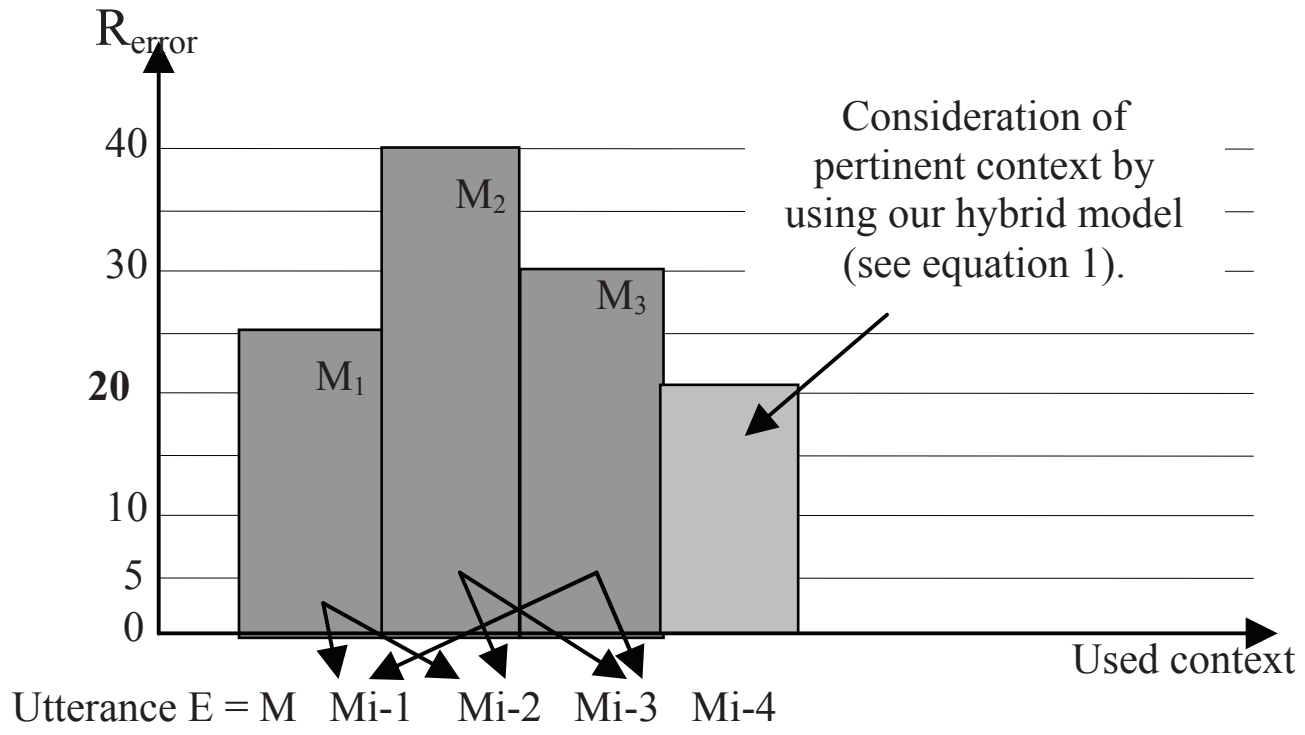

**Fig. 5.** Influence of the pertinent context use on the semantic analyzer performance.

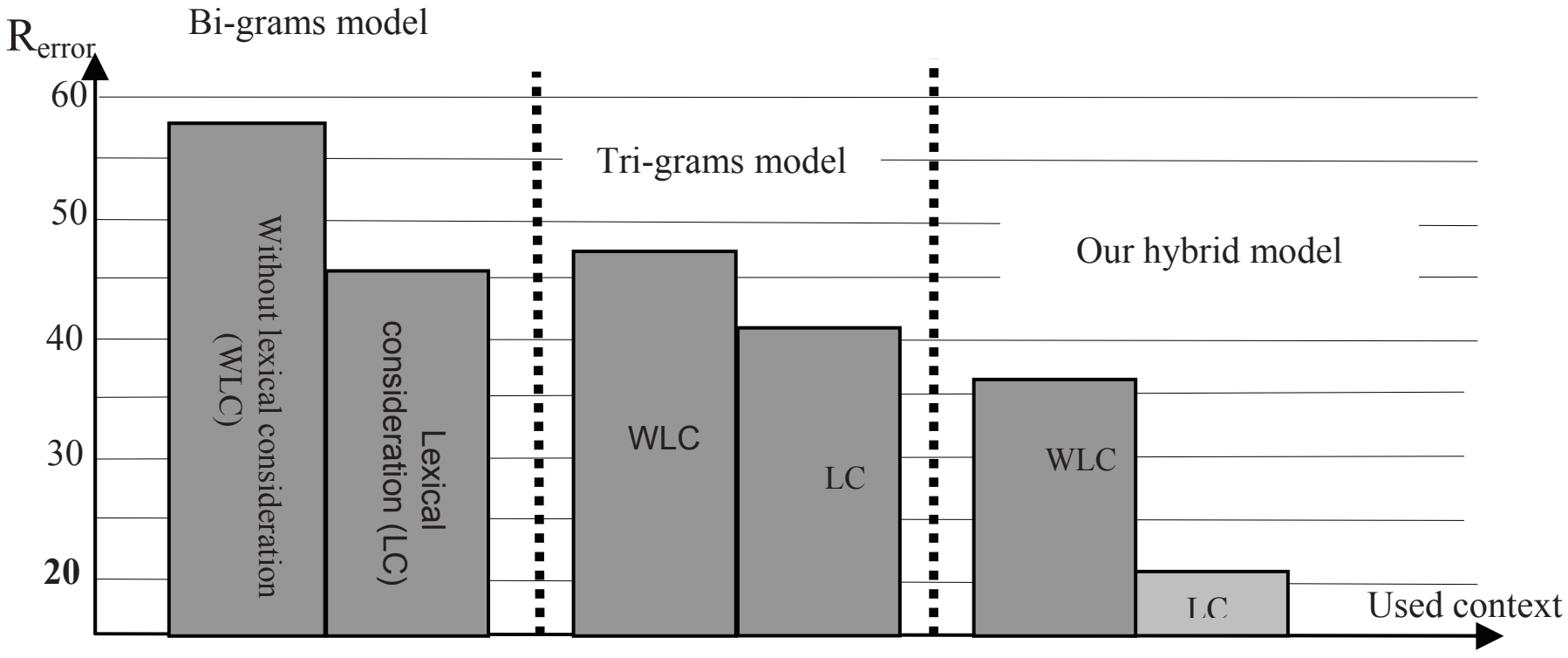

**Fig. 6.** Influence of several types use of the contextual information and of the context length, on the semantic analyzer performance.

The bi-grams and tri-grams models of the above figure 6 are models of Part-of-speech type. Those have allowed us to test the consideration influence of different contexts' length, on the semantic analyzer performance.
According to the above figure 5, it is obvious that the minimum error rate is reached by using only the pertinent contextual informations. In the figure 6, we notice that each time we consider lexical data in a model, the result is improved. The improvement increases by integrating more the utterance type; the error rate reaches 37% (with consideration of the lexical data). By considering later the pertinent FSe (i.e. FSe assigned to the words having strongest affinity with the target word) for the prediction of FSe describing the meaning of the word to be interpreted, we notice that we reach an error rate of about 20%. By analysing our test corpus, we noticed that this error rate is mainly due to the utterances having a very complex syntactic structure. In order to solve this problem, some systems combine a deep syntactic analysis with a selective analysis such as the TINA system of [25]. Other systems use the analyzes strategies of NLP robust [26]. These systems are powerful in open applications.

## 6. Conclusion

In this paper we have presented a semantic analyzer based on a hybrid language model, which helps to integrate simultaneously lexical, semantic and illocutionary contextual data. In addition it not has to take into account only pertinent FSe in the word history. To achieve this goal, we have developed a method based on the average mutual information notion. The results are satisfactory. In the near we have presented future work, we intend to evaluate our model by comparing it to the called distant models or to the models obtained by linear combination of well-known language models like the maximum of entropy. We also hope to define an ungrammaticality gradient to allow evaluating the syntactic complexity of a statement and then choose the adequate model to apply.